\documentclass[12pt]{article}
\usepackage{graphicx}
\usepackage{setspace}
\usepackage{bigints}
\usepackage{graphics} 
\usepackage{epstopdf}
\usepackage{epsfig} 
\usepackage{mathptmx} 
\usepackage{times} 
\usepackage{amsmath} 
\usepackage{amsthm}
\usepackage{amssymb}  
\usepackage{algorithmicx}
\usepackage{algorithm}
\usepackage{bm}
\usepackage[noend]{algpseudocode}
\usepackage{cite}
\usepackage{subfig}
\usepackage{gensymb}
\usepackage{booktabs}
\usepackage{url}
\usepackage{tikz,pgfplots}
\pgfplotsset{compat=newest} 
\pgfplotsset{plot coordinates/math parser=false} 

\DeclareMathOperator*{\minimize}{minimize}

\DeclareMathOperator*{\subjectto}{subject\:to}

\newtheorem{problem}{Problem}

\DeclareMathAlphabet{\pazocal}{OMS}{zplm}{m}{n}

\newlength\figureheight 
\newlength\figurewidth  

\makeatletter
\def\my@tag@font{\normalsize}
\def\maketag@@@#1{\hbox{\m@th\normalfont\my@tag@font#1}}
\let\amsmath@eqref\eqref
\renewcommand\eqref[1]{{\let\my@tag@font\relax\amsmath@eqref{#1}}}
\makeatother

\newcommand{\irow}[1]{
	\left[\begin{smallmatrix}#1\end{smallmatrix}\right]%
}

\title{\LARGE \bf On sensing-aware model predictive path-following control for a reversing general 2-trailer with a car-like tractor} 

\author
{\small Oskar~Ljungqvist$^1$, Daniel~Axehill$^1$, Henrik~Pettersson$^2$\\
	\small{$^1$Department of Automatic Control, Link\"oping University, Link\"oping, Sweden.} \\ 
	\small{E-mail: \texttt{\{oskar.ljungqvist, daniel.axehill\}@liu.se.}} \\
		\small{$^2$ Scania CV, S\"odert\"alje, Sweden.} \\ 
	\small{E-mail: \texttt{henrik\_x.pettersson@scania.com.}}
}

\date{}

\begin{document} 
		
\baselineskip 16pt
	
\maketitle 

\begin{abstract} 
The design of reliable path-following controllers is a key ingredient for successful deployment of self-driving vehicles. This controller-design problem is especially challenging for a general 2-trailer with a car-like tractor due to the vehicle's structurally unstable joint-angle kinematics in backward motion and the car-like tractor's curvature limitations which can cause the vehicle segments to fold and enter a jackknife state. Furthermore, advanced sensors with a limited field of view have been proposed to solve the joint-angle estimation problem online, which introduce additional restrictions on which vehicle states that can be reliably estimated. To incorporate these restrictions at the level of control, a model predictive path-following controller is proposed. By taking the vehicle's physical and sensing limitations into account, it is shown in real-world experiments that the performance of the proposed path-following controller in terms of suppressing disturbances and recovering from non-trivial initial states is significantly improved compared to a previously proposed solution where the constraints have been neglected.       
\end{abstract}

\section{Introduction}
Over the past decades, autonomous transport solutions and advanced driver assistance systems have received a rapidly increased interest as the technology in these areas have advanced. 
Today, autonomous driving in urban areas still faces many unsolved problems and legislation changes are needed to drive autonomously on public roads. 
In contrast, autonomous driving in closed areas such as mines and loading sites are more suitable for initial deployment of such systems.
Within these sites, different tractor-trailer vehicles are frequently used. 
These vehicles are composed of a car-like tractor and $N$ passive trailers that are interconnected though hitches that are of off-axle or on-axle type. When the connections between the vehicle segments are of mixed hitching types, the tractor-trailer vehicle is called a general N-trailer (GNT) and when only pure on-axle or pure off-axle hitching is present, it is referred to as a standard N-trailer (SNT) or a non-standard N-trailer (nSNT), respectively.    
Due to the kinematic properties of tractor-trailer vehicle's~\cite{sordalen1993conversion,altafini1998general,CascadeNtrailernonmin}, the feedback-control problem is in general very difficult. The different feedback-control problems that have been investigated for GNT or nSNT are mainly path following (see e.g.,~\cite{LjungqvistJFR2019,hybridcontrol2001,altafini2003path,rimmer2017implementation,astolfi2004path,Cascade-nSNT,bolzern1998path,wu2017path}), trajectory tracking and set-point stabilization (see e.g.,~\cite{CascadeNtrailer,michalek2018forward,kayacan2014robust}), or similar control problems for the differentially flat SNT vehicle (see e.g.,~\cite{sampei1995arbitrary,SamsonChainedform1995,sordalen1993conversion}).  

\begin{figure}[t]
	\centering
	\includegraphics[width=1\linewidth]{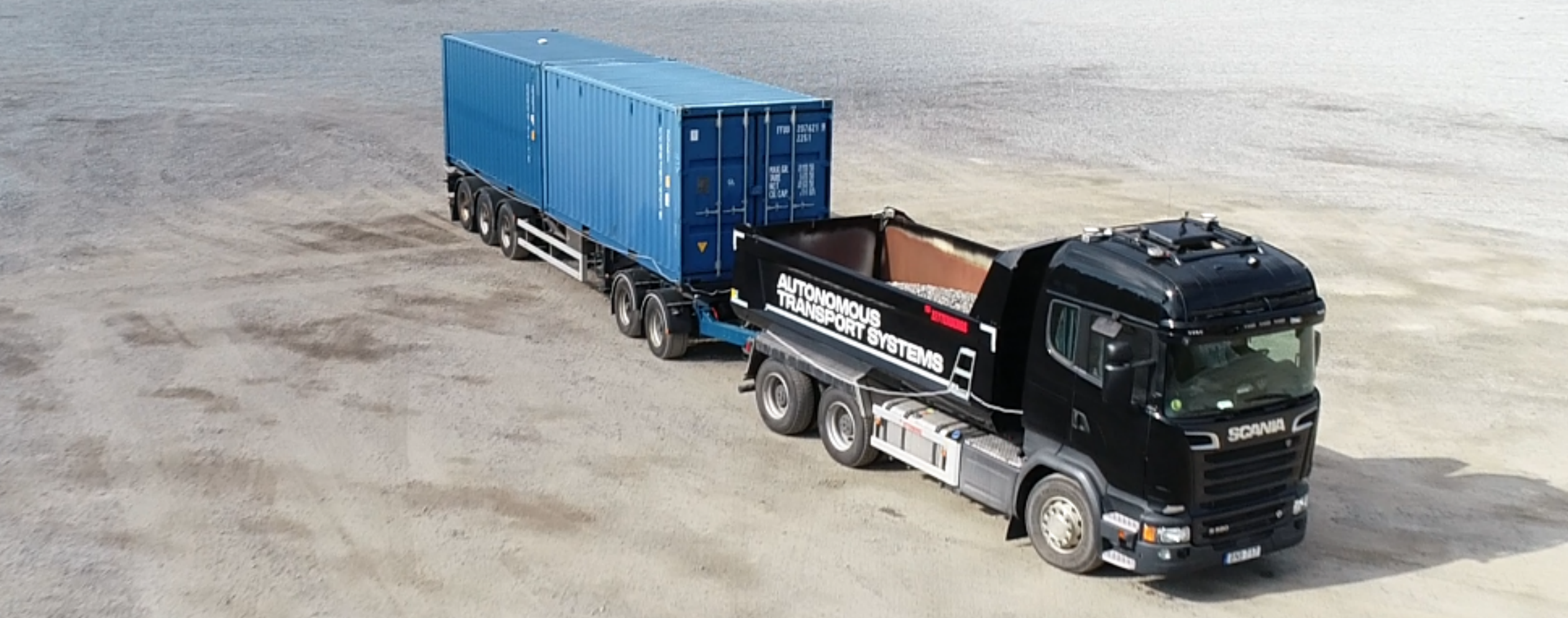}   
	\caption{The test vehicle that is used as research platform. The tractor is a modified version of a Scania R580 6x4 tractor.}	\label{c7:fig:truck_scania} 
\end{figure}

Most of the control approaches presented above consider the problem of following a trajectory or path defined in the position and orientation of the last trailer's axle.
In this work, the path-following control problem during low-speed maneuvers for a G2T with a car-like tractor (see Figure~\ref{c7:fig:truck_scania}) is addressed for the case when the nominal path contains full state and control information, i.e., it is tailored to operate in series with a motion planner as in~\cite{evestedtLjungqvist2016planning,LjungqvistJFR2019,li2019trajectory}. In such an architecture, to avoid collision with surrounding obstacles it is crucial that all nominal vehicle states are followed. The control problem considered in this work is challenging due to the vehicle's structurally unstable joint-angle kinematics in backward motion which can lead to jackknifing\footnote{Jackknifing refers to the situation when the vehicle segments are folding together.}, and the car-like tractor's limited curvature and curvature rate. Additionally, advanced sensors such as cameras or LIDARs mounted in the rear of the car-like tractor have been proposed to solve the joint-angle estimation problem online~\cite{CameraSolSaxe,caup2013video,LjungqvistJFR2019,Daniel2018,Patrik2016}. Since these sensors typically have a limited field of view (FOV), it is important that the vehicle is controlled such that its joint angles remain in the region where high-accuracy state estimates can be reliably computed online. 

The contribution of this work is the extension of the path-following controller presented in our previous work in~\cite{LjungqvistJFR2019}, to explicitly take into account the vehicle's physical and sensing limitations in the controller. It is done by proposing a path-following controller based on the framework of model predictive control (MPC)~\cite{faulwasser2015nonlinear,mayne2000constrained,pedroSpatial}. The proposed MPC controller is evaluated in a set of real-world experiments, where its performance in terms of suppressing disturbances and satisfying the vehicle's physical and sensing imposed constraints is significantly improved compared to the proposed linear quadratic (LQ) controller presented in our previous work in~\cite{LjungqvistJFR2019} where the constraints have been neglected.

The remainder of the paper is organized as follows. In Section~\ref{c7:sec:Modeling}, the path-following error model and the control problem formulation are presented. In Section~\ref{c7:sec:MPC}, the proposed MPC controller and the control design are presented. Results from real-world experiments on a full-scale test vehicle are presented in Section~\ref{c7:sec:results} and the paper is concluded in Section~\ref{c7:sec:conclusions} by summarizing the contributions and a discussion of directions for future work.

\section{Modeling and problem formulation}
\label{c7:sec:Modeling}
\begin{figure}[t!]
	\centering
	\includegraphics[width=0.8\linewidth]{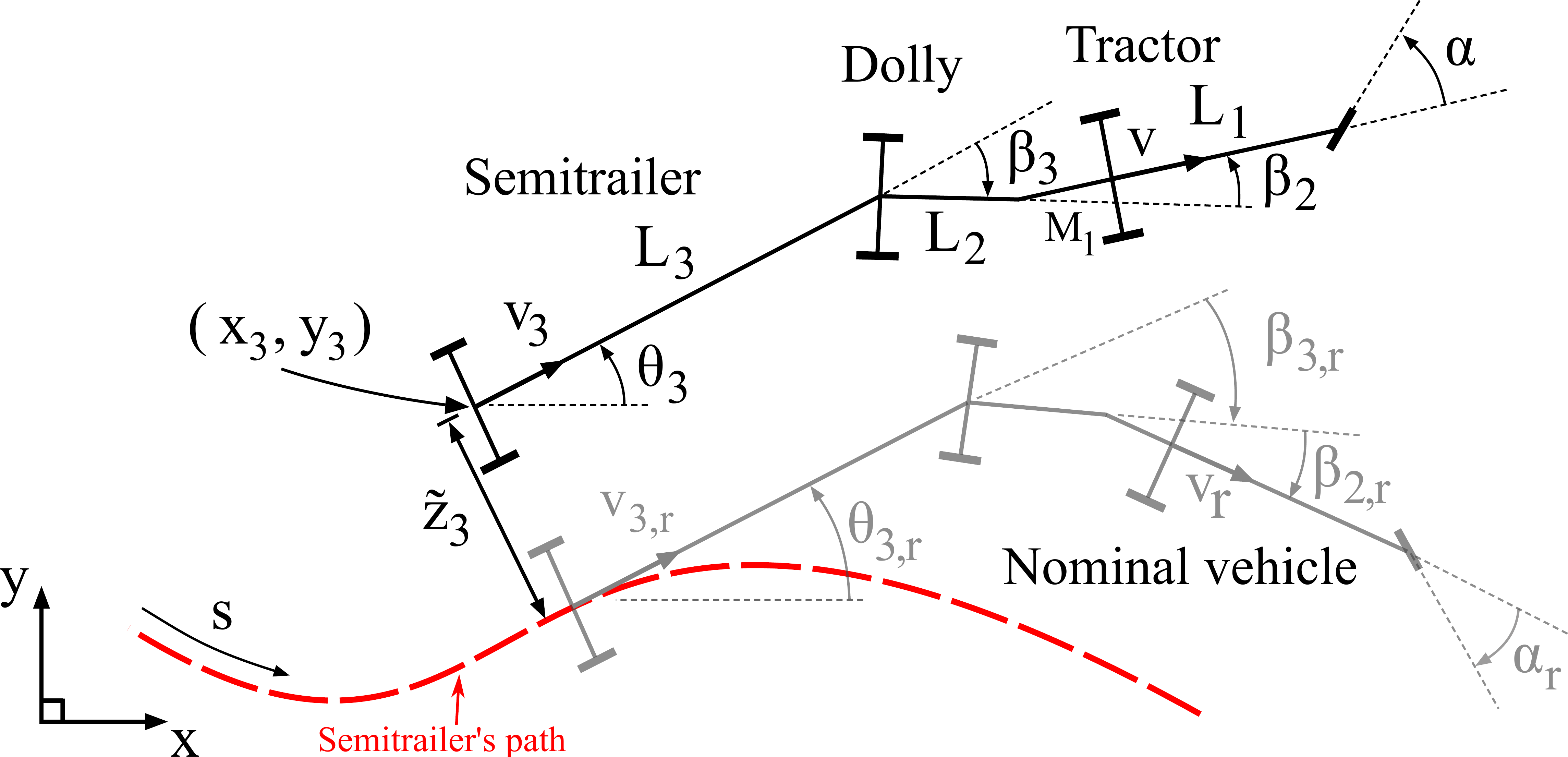}
	\caption{A schematic description of the geometric lengths, states and control signals that are of relevance for modeling the G2T with a car-like tractor in a global and the Frenet frame coordinate system.}
	\label{c7:fig:frenet}%
\end{figure}

The full-scale G2T with a car-like tractor considered in this work (see Figure~\ref{c7:fig:truck_scania}) is schematically illustrated in Figure~\ref{c7:fig:frenet}. 
The tractor-trailer vehicle has a positive off-axle connection between the car-like tractor and the dolly, and an on-axle connection between the dolly and the semitrailer. 
The state vector $x=\irow{x_3&y_3 &\theta_3&\beta_3&\beta_2}^T$ is used to represent a configuration of the vehicle, where $(x_3,y_3)$ denotes the position of the center of the semitrailer's axle, $\theta_3$ is the orientation of the semitrailer, $\beta_3$ is the joint angle between the semitrailer and the dolly, and $\beta_2$ is the joint angle between the dolly and the car-like tractor. 
The length $L_3$ represents the distance between the axle of the semitrailer and the axle of the dolly, $L_2$ is the distance between the axle of the dolly and the off-axle hitching connection at the car-like tractor, $M_1$ is the length of the off-axle hitching (positive in Figure~\ref{c7:fig:frenet}), and $L_1$ denotes the wheelbase of the car-like tractor. 
The car-like tractor is assumed to be front-wheeled steered with perfect Ackermann steering geometry, where $\alpha$ denotes its steering angle. The control signals to the system are the curvature \mbox{$u = \frac{\tan \alpha}{L_1}$} of the car-like tractor and the longitudinal velocity $v$ of its rear axle. 
A recursive formula derived from nonholonomic and holonomic constraints for the GNT vehicle is presented in~\cite{altafini1998general}. Applying the formula for this specific tractor-trailer system results in the following kinematic vehicle model~\cite{hybridcontrol2001}:
\begin{subequations}
	\begin{align} 
	\dot{x}_3 &= v_3 \cos \theta_3,  \label{eq:model1}\\
	\dot{y}_3 &= v_3 \sin \theta_3,  \label{eq:model2} \\
	\dot{\theta}_3 &= v_3 \frac{\tan \beta_3 }{L_3}, \label{eq:model3}\\
	\dot{\beta}_3 &= v_3\left(\frac{\sin\beta_2 - M_1\cos \beta_2u}{L_2  C_1(\beta_2,\beta_3,u)} - \frac{\tan\beta_3}{L_3}\right), \label{eq:model4}\\
	\dot{\beta}_2 &= v_3 \left( \frac{u - \frac{\sin \beta_2}{L_2} + \frac{M_1}{L_2}\cos \beta_2 u}{C_1(\beta_2,\beta_3,u)}\right), \label{eq:model5}
	\end{align}
	\label{c7:eq:model}
\end{subequations}
where $C_1(\beta_2,\beta_3,u)$ is defined as 
\begin{align}
C_1(\beta_2,\beta_3,u) = \cos\beta_3\left(\cos{\beta_2} + M_1\sin\beta_2u\right),
\label{c7:eq:C1}
\end{align}
which describes the relationship, $v_3 = vC_1(\beta_2,\beta_3,u)$, between the velocity $v_3$ of the axle of the semitrailer and the velocity $v$ of the rear axle of the car-like tractor.  When $C_1(\beta_2,\beta_3,u)=0$, the system in~\eqref{c7:eq:model} is singular and not proven controllable~\cite{altafini1998general}, it is therefore further assumed that \mbox{$C_1(\beta_2,\beta_3,u)>0$}. 
Represent the model in~\eqref{c7:eq:model} as $\dot x = v_3f(x,u)$. This model is derived based on no-slip assumptions and the vehicle is assumed to operate on a flat surface. Here, the focus is on low-speed maneuvers and these assumptions are thus expected to hold.

Since time-scaling~\cite{sampei1986time} can be applied to eliminate the velocity dependence in~\eqref{c7:eq:model}, it is without loss of generality further assumed that the velocity of the tractor $v\in\{-1,1\}$. The direction of motion is essential for the stability of the system~\eqref{c7:eq:model}, where the joint angles are structurally unstable in backward motion ($v < 0$), where it risks enter what is called a jack-knife state~\cite{hybridcontrol2001}. 
In forward motion \mbox{($v > 0$)}, these modes are stable but due to the positive off-axle hitching ($M_1>0$), the system possess non-minimum phase properties in some of its output channels~\cite{CascadeNtrailernonmin}.

\subsection{Constraints}
The car-like tractor is assumed to have physical bounds on its curvature $u$ and curvature rate $\dot u$, which are modeled using box constraints
\begin{align}
\label{c7:control_constraints}
|u|\leq u_{\text{max}},\quad |\dot u|\leq \dot u_{\text{max}},
\end{align} 
where the positive constants $u_{\text{max}}$ and $\dot u_{\text{max}}$ denote maximum curvature and curvature rate, respectively. In practice, advanced sensors mounted on the car-like tractor are commonly used for online joint-angle estimation~\cite{CameraSolSaxe,caup2013video,LjungqvistJFR2019,Daniel2018,Patrik2016}. 
Such sensors typically have a limited FOV which enforce restrictions on the joint angles that can be estimated with high accuracy. Additionally, constraints on the joint angles that prevent the vehicle to enter a jackknife state should also be considered. The above mentioned restrictions are modeled by a convex polytope:
\begin{align}
\label{c7:joint_angle_constaints}
\left(\beta_3,\beta_2\right)\in\mathbb P =  \left\{(\beta_3,\beta_2)\in\mathbb R^2 \middle| H\left(\begin{matrix}
\beta_3 & \beta_2
\end{matrix}\right)^T \leq h \right\} ,
\end{align}
where $H\in\mathbb R^{m \times 2}$, $h\in\mathbb R^{m}$ and $m\in\mathbb Z_{+}$. Even though the constraints in~\eqref{c7:control_constraints} and \eqref{c7:joint_angle_constaints} have been consider by the motion planner, disturbances are always present during plan execution, which makes it important to also explicitly consider the them at the level of control.

\subsection{Path-following error model}
Given a nominal trajectory $(x_r(\cdot),u_r(\cdot),v_{3r}(\cdot))$ satisfying the model of the G2T with a car-like tractor~\eqref{c7:eq:model}:
\begin{align}
\dot{x}_r = v_{3r}f( x_r,u_r), \label{c7:eq:tray}
\end{align}
which is assumed to satisfy the constraints in~\eqref{c7:control_constraints} and~\eqref{c7:joint_angle_constaints}. 
When the non-time critical path-following control problem is considered, it is convenient to separate the longitudinal from the lateral control of the vehicle~\cite{reviewFrazzoli2016}. Except from satisfying the constraints, the objective of the lateral controller, i.e., the path-following controller, is to control the tractor's curvature $u$ such that the motion plan obtained from the motion planner is traversed with a small and bounded path-following error. 
In analogy to~\cite{LjungqvistJFR2019}, this is done by first deriving a path-following error model. 
Introduce $s(t)$ as the distance traveled by the axle of the semitrailer onto its projection to its nominal path $(x_{3r}(\cdot), y_{3r}(\cdot))$ up to time $t$. Since $\frac{\text dx_r}{\text dt}=\frac{\text dx_r}{\text ds}|v_{3r}|$, the nominal trajectory in~\eqref{c7:eq:tray} can instead be parametrized in $s$ and interpreted as a nominal path
\begin{align}
\frac{\text d x_r}{\text ds} = \bar v_{3r}f(x_r,u_r), \label{c7:eq:tray_s}
\end{align}
where $\bar v_{3r}=\text{sign}(v_{3r})\in\{-1,1\}$ specifies the nominal motion direction, i.e., $\bar v_{3r}=1$ and $\bar v_{3r}=-1$ represent forward and backward motion, respectively. To exploit that a nominal path~\eqref{c7:eq:tray_s} is provided, the G2T with car-like tractor in~\eqref{c7:eq:model} is first described in terms of deviation from the nominal path generated by the system in \eqref{c7:eq:tray_s}, see Figure~\ref{c7:fig:frenet}.
Denote $\tilde z_3(t)$ as the signed lateral distance between the center of the semitrailer's axle onto its projection to its nominal path $(x_{3r}(\cdot), y_{3r}(\cdot))$. 
Let $\tilde{\theta}_3(t)=\theta_3(t)-\theta_{3r}(s(t))$ be the orientation error of the semitrailer, while $\tilde{\beta}_3(t)=\beta_3(t)-\beta_{3r}(s(t))$ and $\tilde{\beta}_2(t)=\beta_2(t)-\beta_{2r}(s(t))$ are the joint-angle errors. Finally, denote the deviation in the tractor's curvature as $\tilde{u}(t) =u(t) - u_r(s(t))$.
Using the Frenet-frame transformation together with the chain rule (see~\cite{LjungqvistJFR2019} for details), the controlled vehicle~\eqref{c7:eq:model} can be described in terms of deviation from the nominal path~\eqref{c7:eq:tray_s} as
\begingroup\makeatletter\def\f@size{10}\check@mathfonts
\begin{subequations}
	\begin{align}
	\dot s &= v_3 \frac{\bar v_{3r}\cos \tilde \theta_3}{1-\kappa_{3r} \tilde z_3}, 
	\label{c7:eq:model_s1}
	\\ 
	\dot{\tilde z}_3 &= v_3 \sin \tilde \theta_3, 
	\label{c7:eq:model_s2}
	\\
	\dot{\tilde \theta}_3 &= v_3 \left( \frac{\tan(\tilde{\beta}_3+\beta_{3r})}{L_3} - \frac{\kappa_{3r}\cos \tilde \theta_3}{1-\kappa_{3r} \tilde z_3} \right), \label{c7:eq:model_s3}
	\\
	\dot{\tilde \beta}_3 &= v_3\left(\frac{\sin(\tilde \beta_2+\beta_{2r})-M_1\cos(\tilde \beta_2+\beta_{2r})(\tilde u+ u_r)}{L_2 C_1(\tilde \beta_2+\beta_{2r},\tilde \beta_3+\beta_{3r}, \tilde u + u_r)} - \frac{\tan(\tilde \beta_3+\beta_{3r})}{L_3}  -\frac{\cos{\tilde{\theta}_3}}{1-\kappa_{3r}\tilde z_3}\left[\frac{\sin\beta_{2r} - M_1\cos\beta_{2r} u_r}{L_2C_1(\beta_{2r},\beta_{3r},u_r)}-\kappa_{3r}\right]\right), 
	\label{c7:eq:model_s4} \\
	\dot{\tilde \beta}_2 &=v_3\left( \left[ \frac{\tilde u+ u_r - \frac{\sin(\tilde \beta_2+\beta_{2r})}{L_2} + \frac{M_1}{L_2}\cos(\tilde \beta_2+\beta_{2r})(\tilde u+ u_r)}{C_1(\tilde \beta_2+\beta_{2r},\beta_3+\beta_{3r}, \tilde u+ u_r)}\right]  -\frac{\cos{\tilde{\theta}_3}}{1-\kappa_{3r}\tilde z_3}\left[ \frac{u_r - \frac{\sin \beta_{2r}}{L_2} + \frac{M_1}{ L_2}\cos \beta_{2r}u_r}{C_1(\beta_{2r},\beta_{3r}, u_r)}\right]\right), \label{c7:eq:model_s5}
	\end{align}
	\label{c7:eq:model_frenet_frame}
\end{subequations}
\endgroup
where
\begin{equation}
\kappa_{3r}(s)=\frac{\text d\theta_{3r}}{\text ds}=\frac{\tan \beta_{3r}(s)}{L_3},
\end{equation}
is the nominal curvature for the center of the axle of the semitrailer. The transformation to the Frenet frame path-coordinate system is valid as long as $\tilde z_3$ and $\tilde\theta_3$ satisfy 
\begin{align}\label{c7:frenet_frame_transformation_constraints}
1-\kappa_{3r}(s)\tilde z_3>0, \quad |\tilde\theta_3|<\pi/2.
\end{align}
Essentially, we must have that $|\tilde z_3|<|\kappa_{3r}^{-1}(s)|$ when $\tilde z_3$ and $\kappa_{3r}(s)$ have the same sign~\cite{LjungqvistJFR2019}.
Furthermore, $\bar v_{3r}$ is included in~\eqref{c7:eq:model_s1} to make $\dot s> 0$ as long as the constraints in~\eqref{c7:frenet_frame_transformation_constraints} are satisfied, and the semitrailer's velocity $v_3$ and the nominal direction $\bar v_{3r}$ have the same sign. 
Since it is assumed that $C_1(\beta_2,\beta_3,u)>0$ and the relationship $v = v_3C_1(\beta_2,\beta_3,u)$ holds, this is equivalent to that the velocity of the rear axle of the car-like tractor $v$ is selected such that is has the same sign as the nominal motion direction $\bar v_{3r}$. 
\subsection{Problem formulation}
Define the path-following error state $\tilde x = \irow{\tilde z_3 & \tilde\theta_3 & \tilde\beta_3 & \tilde\beta_2}^T$, where its model is given by~\eqref{c7:eq:model_s2}--\eqref{c7:eq:model_s5}. It is easily verified that the origin $(\tilde x,\tilde u)=(0,0)$ to this system is an equilibrium point for all $t$. Formally, the following control problem is considered:
\begin{problem}
	\label{c7:path_following_contr_problem}
	Given the path-following error model in~\eqref{c7:eq:model_s2}--\eqref{c7:eq:model_s5}, design a feedback control law $\tilde u(t)=g(s(t),\tilde x(t))$ which achieves
	\begin{enumerate}
		\item[i)] \textit{Path convergence}: The path-following error states $\tilde x(t)$ converges towards the origin in the sense that
		\begin{align*}
		\lim\limits_{t\rightarrow\infty} \lVert \tilde x (t) \rVert = 0.
		\end{align*} 
		\item[ii)] \textit{Constraint satisfaction}: For all $t\in[0,\infty)$, the constraints on the joint angles $(\beta_2,\beta_3)\in\mathbb P$, and controlled curvature $|u(t)|\leq u_{\text{max}}$ and curvature rate $|\dot u(t)|\leq \dot u_{\text{max}}$ are satisfied.  
	\end{enumerate}
\end{problem}
This problem formulation is closely related to the one proposed in~\cite{faulwasser2015nonlinear} if $s(t)$ is viewed as the so called \textit{timing-law}. However, in this setup, the progression along the nominal path $s(t)$ is not explicitly controlled by the path-following controller, as it is determined by the projection of the semitrailer's position onto its nominal path.

Since the velocity of the tractor $v$ is selected such that $\dot s(t)>0$, it is possible to perform time-scaling~\cite{sampei1986time} and eliminate the time-dependency presented in~\eqref{c7:eq:model_s2}--\eqref{c7:eq:model_s5}. Using the chain rule, it holds that $\frac{\text d\tilde x}{\text  ds}=\frac{\text d\tilde x}{\text  dt}\frac{1}{\dot s}$, and the distance-based version of the path-following error model~\eqref{c7:eq:model_s2}--\eqref{c7:eq:model_s5} can compactly be written as
\begin{align}
\frac{\text d\tilde x}{\text ds}=\tilde f(s,\tilde x,\tilde u),
\label{c7:eq:MPC_spatial_path_following_error_model}
\end{align} 
where the origin $(\tilde x,\tilde u)=(0,0)$ is an equilibrium point for all $s$. In the next section, a solution to Problem~\ref{c7:path_following_contr_problem} is proposed by designing a model predictive path-following controller.   

\section{Model predictive path-following controller}
\label{c7:sec:MPC}
The task of the MPC controller is to control the tractor's curvature $u$ such that the path-following error is minimized while the vehicle's constraints are satisfied for all time instances. To obtain an MPC problem that has potential of being solved at a sufficiently high sampling rate, the goal is to derive an MPC problem formulation that can be converted into the form of a quadratic programming (QP) problem. First, the nonlinear path-following error model~\eqref{c7:eq:MPC_spatial_path_following_error_model} is linearized around the origin $(\tilde x,\tilde u) = (0,0)$:
\begin{align}
\label{c7:linear_cont_model}
\frac{\text d \tilde x}{\text ds} = A(s)\tilde x +  B(s)\tilde u.
\end{align}
With the sampling distance $\Delta_s$, Euler-forward discretization yields a discrete approximation of~\eqref{c7:linear_cont_model} in the form
\begin{align}
\label{c7:d_lin_sys}
\tilde x_{k+1} = F_k\tilde x_k + G_k\tilde u_k,
\end{align}  
where
\begin{align}
\label{c7:d_system_matrix}
F_k = I +\Delta_s A_k, \quad
G_k = \Delta_sB_k.
\end{align}    
Since the tractor's curvature \mbox{$u_k=\tilde u_k+u_{r,k}$}, the deviation in the curvature is bounded as
\begin{align}
\label{c7:curvature_constaint_MPC}
-u_{\text{max}} \leq \tilde u_k + u_{r,k}  \leq u_{\text{max}}.
\end{align}  
Furthermore, since $\dot s>0$ the tractor's limited curvature rate $|\dot u|\leq \dot u_{\text{max}}$ can be described in $s$ using the chain rule as
\begin{align}
\label{c7:rate-limit-constraints-ds}
\left|\frac{\text du}{\text ds}\right|\leq\frac{\dot u_{\text{max}}}{\dot s} = \frac{1-\kappa_{3r} \tilde z_3}{|v| C_1(\beta_2,\beta_3,u)\cos \tilde \theta_3}\dot u_{\text{max}},
\end{align}
since $v_3 = v C_1(\beta_2,\beta_3,u)$. Locally around the nominal path $(\tilde x,\tilde u)=(0,0)$, it holds that $\cos\tilde\theta_3\approx 1$ and $\kappa_{3r} \tilde z_3\approx 0$. Thus, to avoid coupling between $\tilde x$ and $\tilde u$, the constraint in~\eqref{c7:rate-limit-constraints-ds} is approximated as
\begin{align}
\label{c7:rate-limit-constraints-ds2}
\left|\frac{\text du}{\text ds}\right| \leq \frac{\dot u_{\text{max}}}{|v|C_1(\beta_{2r}(s),\beta_{3r}(s),u_r(s))} \triangleq c_{\text{max}}(s).
\end{align} 
By discretizing~\eqref{c7:rate-limit-constraints-ds2} using Euler forward, the rate limit on the controlled curvature deviation can be described by the following slew-rate constraint
\begin{align}
\label{c7:curvature_rate_constaint_MPC}
-c_{\text{max},k}\Delta_s \leq \tilde u_{k} - \tilde u_{k-1} - \bar u_{r,k} \leq c_{\text{max},k}\Delta_s,   
\end{align}
where $\bar u_{r,k} = u_{r,k} - u_{r,k-1}$. Denote the linear inequality constraints in~\eqref{c7:curvature_constaint_MPC} and~\eqref{c7:curvature_rate_constaint_MPC} as \mbox{$\tilde u_k\in\tilde {\mathbb U}_k$}. 
Finally, since \mbox{$\beta_{j,k}=\beta_{jr,k}+\tilde\beta_{j,k}$} for $j=2,3$, the convex polytope $\mathbb P$ in~\eqref{c7:joint_angle_constaints} that represents the constraint on the joint angles can be written as
\begin{align}
\label{c7:joint_angle_constaints_error_states}
H\left(\begin{matrix}
\tilde\beta_{3,k} \\ \tilde\beta_{2,k}
\end{matrix}\right) \leq h - H\left(\begin{matrix}
\beta_{3r,k} \\ \beta_{2r,k}
\end{matrix}\right) \triangleq \bar h_{k},
\end{align}
which is denoted by $(\tilde\beta_{3,k},\tilde\beta_{2,k})\in\tilde{\mathbb P}_{k}$. Now, assume that full-state information $\tilde x(s(t))$ at time $t$ (or distance $s$) is provided. Then, the MPC problem with prediction horizon $N$ is defined as follows
\begin{subequations}
	\label{c7:MPC_problem}
	\begin{align}
	\minimize_{\mathbf{\tilde x},\hspace{0.1ex} \bm{\tilde u}} \hspace{2ex} &V_N(\mathbf{\tilde x},\bm{\tilde u}) = V_f(\tilde x_N) + \sum_{k=0}^{N-1} l(\tilde x_k,\tilde u_k) \\
	\subjectto \hspace{1ex} 
	&\tilde x_{k+1} = F_k\tilde x_k + G_k\tilde u_k, \\
	& (\tilde\beta_{3,k},\tilde\beta_{2,k})\in \tilde{\mathbb P}_{k}, 
	\quad \tilde u_k\in\tilde{\mathbb U}_k, \\
	&\tilde x_0  = \tilde x(s(t))  \text{ given,}
	\end{align}
\end{subequations}
for $k = 0,1,\hdots,N-1$, where $\mathbf{\tilde x}^T = \irow{\tilde x_0^T & \tilde x_1^T & \hdots & \tilde x_N^T}$ is the predicted path-following error state-vector sequence and \mbox{$\mathbf{\tilde u} = \irow{\tilde u_0 & \tilde u_1 & \hdots &\tilde u_{N-1}}^T$} is the curvature deviation sequence. The stage-cost is chosen to be quadratic $l(\tilde x_k,\tilde u_k) = ||\tilde x_k||^2_Q + \tilde u_k^2$ and the terminal-cost $V_f(\tilde x_N)=||\tilde x_N||^2_{P_N}$, where $Q\succ 0$ and $P_N\succ 0$ are design matrices. 
Since the cost function $V_N$ is quadratic and there are only linear equality and inequality constraints, the MPC problem~\eqref{c7:MPC_problem} can easily be converted into a QP problem. Thus, at each sampling instance, the QP problem in~\eqref{c7:MPC_problem} is solved to obtain the optimal open-loop controlled curvature deviation sequence $\bm{\tilde u}^*$. As in standard receding horizon control, only the first control signal $\tilde u_0^*$ is deployed to the vehicle
\begin{align}
u(t) = u_r(s(t)) + \tilde u_0^*, 
\end{align}
and the optimizing problem~\eqref{c7:MPC_problem} is then repeatedly solved at each sampling instance using new state information.


\begin{figure}[t]
	\centering
	\includegraphics[width=0.9\linewidth]{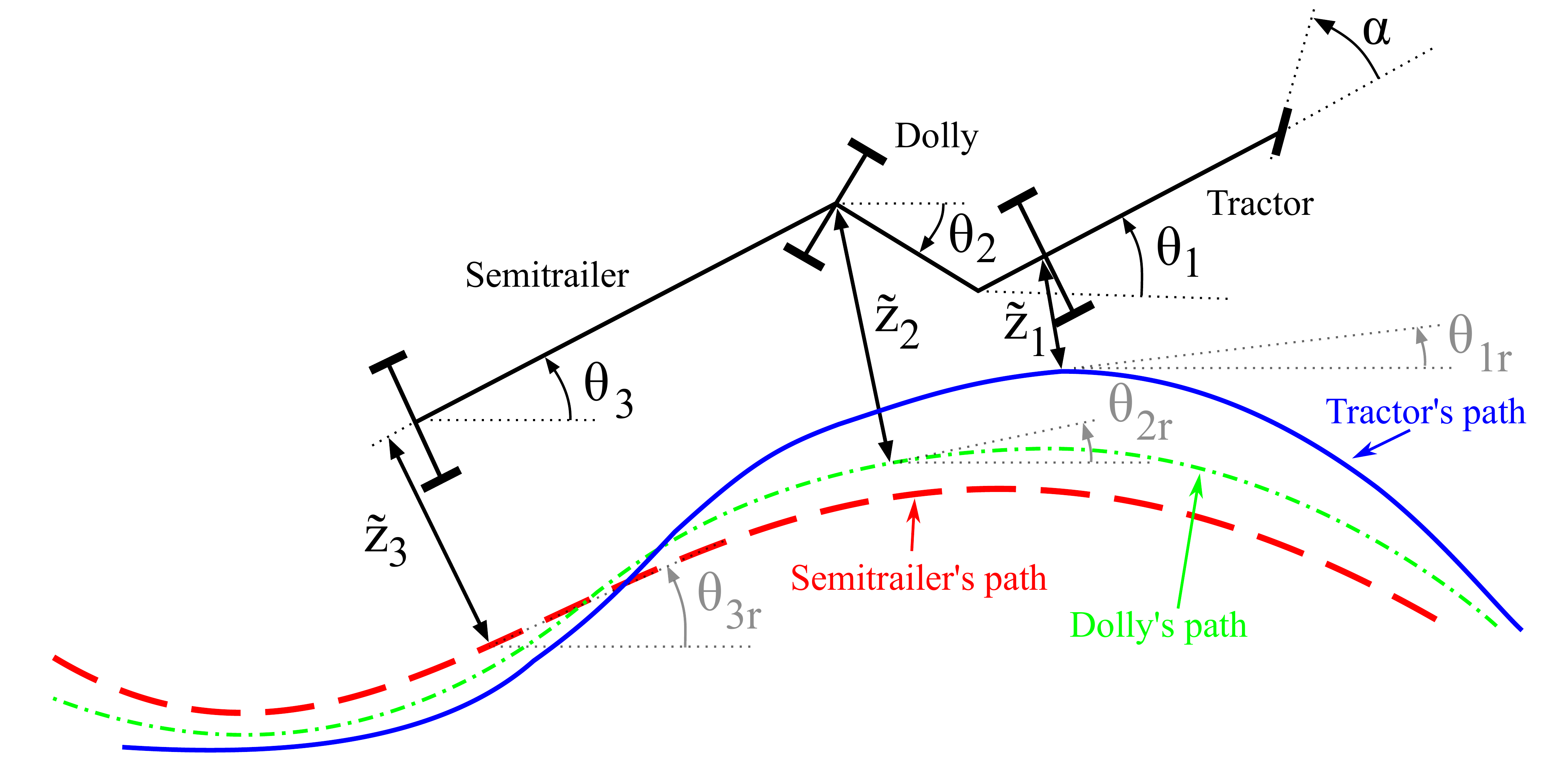}
	\caption{Illustration of the additional path-following error states that are used in the design of the MPC controller.}
	\label{c7:fig:truck_frenet_advanced}
\end{figure}

\subsection{Design of the cost function}
Since a nominal path that satisfies the vehicle model~\eqref{c7:eq:tray_s} is provided, it is possible to calculate the nominal paths for the position and orientation of the dolly as well as the car-like tractor using the nominal state path $x_r(\cdot)$ together with holonomic relationships~\cite{altafini1998general}. To reduce the risk of colliding with surrounding obstacles, the MPC controller should be tuned such that the transient response of all path-following errors are prioritized. Denote $\tilde z_1$ and $\tilde z_2$ as the signed lateral distances of the axle of the dolly and the car-like tractor onto their nominal paths, see Figure~\ref{c7:fig:truck_frenet_advanced}. Moreover, define their corresponding orientation errors as $\tilde \theta_1=\theta_1 -\theta_{1r}$ and $\tilde \theta_2 = \theta_2 -\theta_{2r}$, respectively. In general, it is not possible to derive a closed-form expression to describe these additional path-following error states as a function of the modeled path-following error states $\tilde x$ (see~\cite{altafini2003path} for details). However, for the special case of a straight nominal path, closed-form expressions exist and the signed lateral errors $\tilde z_2$ and $\tilde z_1$ can be described as 
\begin{subequations}
	\label{c7:eq:add_lateral_errors}
	\begin{align}
	\tilde z_2 &= \tilde z_3 + L_3\sin\tilde\theta_3, \\
	\tilde z_1 &= \tilde z_2 + L_2\sin(\tilde\theta_3+\tilde\beta_3) + M_1\sin(\tilde\theta_3+\tilde\beta_3+\tilde\beta_2)
	\end{align}
\end{subequations}
and their corresponding heading errors $\tilde \theta_2$ and $\tilde \theta_1$ as
\begin{subequations}
	\begin{align}
	\tilde \theta_2 &= \tilde\theta_3+\tilde\beta_3, \\
	\tilde \theta_1 &= \tilde\theta_3+\tilde\beta_3+\tilde\beta_2.
	\end{align}
\end{subequations}
Using these approximate relationships, the control-measure vector is defined as 
\begin{align}
z=\left(\begin{matrix}
\tilde z_1 & \tilde\theta_1 & \tilde z_2 & \tilde\theta_2 & \tilde\beta_2 & \tilde z_3 & \tilde\theta_3 & \tilde\beta_3\end{matrix}\right)^T\triangleq h_z(\tilde x).
\end{align}
The function $h_z(\tilde x)$ contains nonlinear terms and its Jacobian linearization around the origin yields $z= \frac{\partial h_z(0)}{\partial \tilde x}\tilde x \triangleq M\tilde x$. By selecting the weight matrix for the quadratic stage-cost as $Q=M^T\bar QM$, where $\bar Q\succeq 0$ is a diagonal matrix, each diagonal element in $\bar Q$ corresponds to penalizing a specific control measure in $z$. The matrix $M$ then transforms a selected design choice to $Q$, which will typically have nonzero off-diagonal elements. 
After $Q$ has been selected, the weight matrix $P_N\succ 0$ for the terminal cost is chosen as the solution to the discrete-time algebraic Riccati equation (DARE):
\begin{align}
\label{c7:eq:DARE}
F^T P_N F - P_N - F^T P_N GK + Q = 0,
\end{align}  
where $K = (1+G^TP_NG)^{-1}G^TP_N F$ is the LQ feedback gain, and the matrices $F=I+\Delta_s A$ and $G=\Delta_s B$ are the discrete system matrices~\eqref{c7:d_system_matrix} for the linearized path-following error model~\eqref{c7:d_lin_sys} around a straight nominal path.

\begin{figure}[t!]
	\centering
	\includegraphics[width=0.7\linewidth]{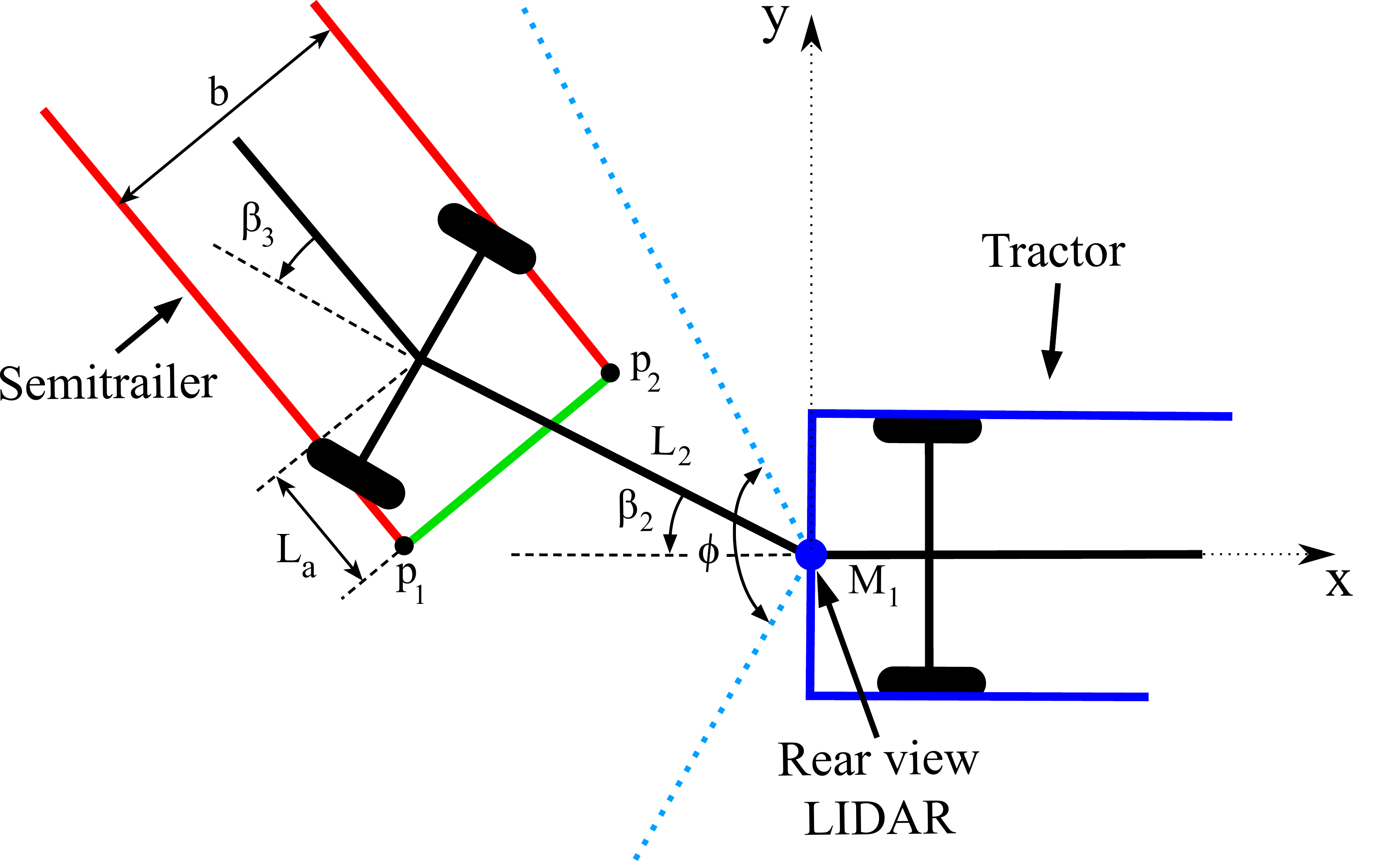}
	\caption{Illustration of the placement of the rear view LIDAR sensor, its FOV (blue dotted lines) and relevant physical quantities. }
	\label{c7:fig:LIDAR_FOV_illustration}
\end{figure}

\subsection{Design of joint-angle constraints} 
The constraint on the joint angles~\eqref{c7:joint_angle_constaints} is intended to be selected such that the system avoids jackknifing, but also to restrict the joint angles to remain in the region where the used state-estimation solution is able to compute reliable and accurate estimates of the vehicle states needed for control.

\begin{figure}[t!]
	\centering
	\includegraphics[width=0.5\linewidth]{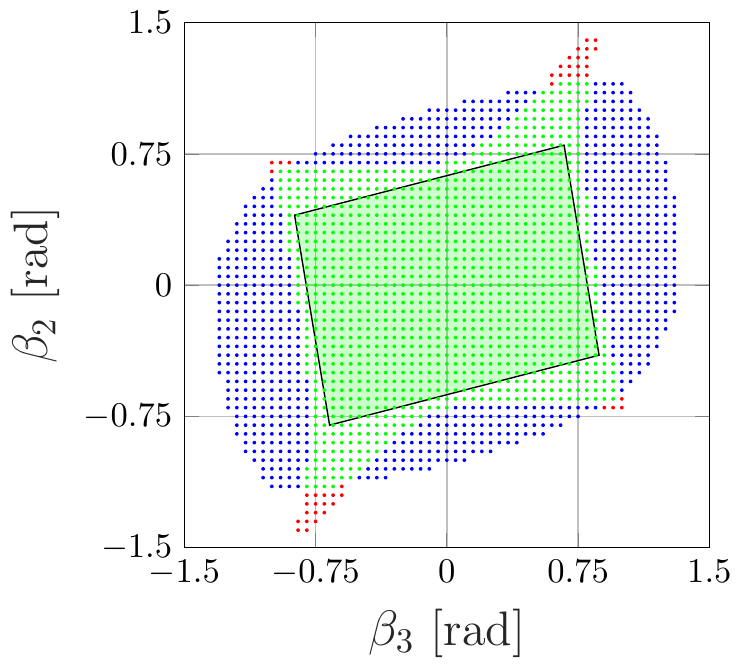}
	\caption{The simulated stability region for different joint angles (blue and green dots) around a straight nominal path and the region where the used LIDAR-based estimation solution can compute estimates of the joint angles (red and green dots). The green polygon is the modeled constraint for the joint angles~\eqref{c7:joint_angle_constaints}.}
	\label{c7:fig:LIDAR_FOV_stability_region}
\end{figure}

Due to the constraints on the tractor's curvature, it is not possible to globally stabilize the path-following error system~\eqref{c7:eq:MPC_spatial_path_following_error_model} since for sufficiently large joint angles, jackknifing is impossible to prevent by only driving backwards~\cite{hybridcontrol2001}. This limit is only possible to calculate analytically for the single-trailer case and approximate methods have to be utilized when more than one trailer is present~\cite{hybridcontrol2001}. Given a straight nominal path, the vehicle parameters presented in Table~\ref{c7:tab:vehicle_parameters}, the MPC controller in~\eqref{c7:MPC_problem} with no joint-angle constraints and the design parameters in Table~\ref{c7:tab:design_parameters}. The system is simulated from different initial joint angles in backward motion with $v=-1$ m/s and is checked for convergence onto the straight nominal path. The resulting simulated stability region for the system is illustrated by the blue and green dots in Figure~\ref{c7:fig:LIDAR_FOV_stability_region}.

What remains is to compute the region where the joint angles can be accurately estimated by the used LIDAR-based estimation solution presented in our previous works in~\cite{LjungqvistJFR2019,Patrik2016,Daniel2018}. The placement of the LIDAR sensor is illustrated in Figure~\ref{c7:fig:LIDAR_FOV_illustration} together with a definition of relevant physical quantities which are also explained in Table~\ref{c7:tab:vehicle_parameters}. Essentially, as long as the entire front of the semitrailer's body (green line in Figure~\ref{c7:fig:LIDAR_FOV_illustration}) is visible from the LIDAR's point cloud, the LIDAR-based estimation solution computes accurate estimates of the joint angles as well as the remaining states of the semitrailer. Using standard trigonometry, this sensing region can be computed numerically. Given the vehicle parameters in Table~\ref{c7:tab:vehicle_parameters}, the computed sensing region is illustrated by the green and red dots in Figure~\ref{c7:fig:LIDAR_FOV_stability_region}. As can be seen, the simulated stability region for the closed-loop system is almost completely covering the system's sensing region. It can therefore be concluded that practically the system is more limited by the sensor properties that the controller performance.

The polytope $\mathbb P$ used to model the constraint on the joint angles~\eqref{c7:joint_angle_constaints} is now selected as an inner approximation of the intersection of these two regions (green polytope in Figure~\ref{c7:fig:LIDAR_FOV_stability_region}). This set is selected conservatively to account for potentially imperfect and delayed joint-angle estimates online. Additionally, since the vehicle model used in the MPC controller only is an approximation, the hard constraints on the joint angles are replaced with soft constraints using slack variables which are added with linear and quadratic penalty to the objective function.

\begin{table}[b!]
	\centering
	\caption{Vehicle parameters.}
	\begin{tabular}{l l}
		\hline \noalign{\smallskip} Vehicle parameter  & Value   \\  \hline \noalign{\smallskip}	
		The tractor's wheelbase $L_1$            &   4.62 m  \\ 
		Maximum curvature $u_{\text{max}}$ & $0.18$ m$^{-1}$ \\
		Maximum curvature rate $\dot u_{\text{max}}$ & $0.13$ m$^{-1}$s$^{-1}$ \\
		Length of the off-hitch $M_1$      &   1.66 m  \\  
		Length of the dolly $L_2$          &   3.87 m  \\    
		Length of the semitrailer $L_3$    &   8.00 m  \\
		Length of the semitrailer's overhang $L_a$       &   1.73 m  \\
		Width of the semitrailer's front $b$   &   2.45 m  \\
		Angle of the horizontal scan field for the LIDAR $\phi$  & $140\times\pi/180$ rad \\
		\hline \noalign{\smallskip}
	\end{tabular}
	\label{c7:tab:vehicle_parameters}
	\vspace{-5pt}
\end{table}

\begin{table}[b!]
	\centering
	\caption{Design parameters.}
	\begin{tabular}{l l}
		\hline \noalign{\smallskip} Design parameter & Value   \\  \hline \noalign{\smallskip}	
		Prediction horizon $N$       &50 \\ 
		Weight matrix $\bar Q$       &$1/35\times$diag$([0.5, 1, 0.5, 1, 4, 0.5, 1,4])$ \\  
		Sampling distance $\Delta_s$ &0.2 m\\
		Controller frequency  $f_s$  &20 Hz \\
		\hline \noalign{\smallskip}
	\end{tabular}
	\label{c7:tab:design_parameters}
\end{table}


\begin{figure}[t!]
	\centering
	\setlength\figureheight{0.5\textwidth}
	\setlength\figurewidth{0.75\textwidth}
	\captionsetup[subfloat]{captionskip=-2pt}
	\subfloat[][The paths taken by the axle of the semitrailer $(x_3(\cdot),y_3(\cdot))$ and the nominal path for $(x_{3r}(\cdot),y_{3r}(\cdot))$ (black solid line).]{
		\begin{tikzpicture}
		\node[anchor=south west] (myplot) at (0,0) {
			\input{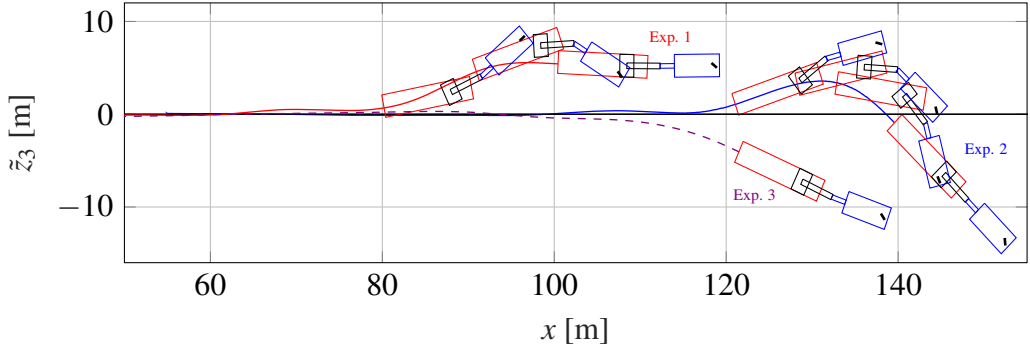}
		};
		\begin{scope}[x={(myplot.south east)}, y={(myplot.north west)}]
		\node[text=red] at (0.65,0.87) {\tiny Exp. 1};
		\node[text=blue] at (0.95,0.57) {\tiny Exp. 2};
		\node[text=red!50!blue] at (0.73,0.45) {\tiny Exp. 3};
		\end{scope}
		\end{tikzpicture}
		\label{c7:fig:traj_straight_xy}
	}
	\quad
	\setlength\figureheight{0.18\textwidth}
	\setlength\figurewidth{0.37\textwidth}
	\subfloat[][The curvature of the car-like tractor.]{
		\begin{tikzpicture}
		\node[anchor=south west] (myplot) at (0,0) {
			\input{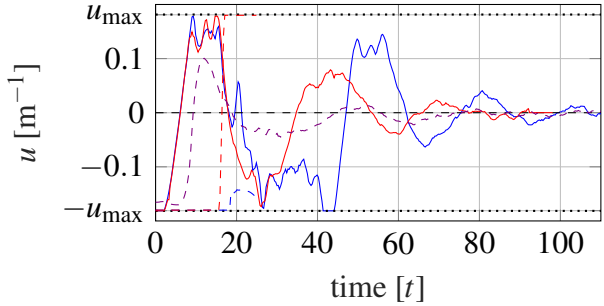}
		};
		\end{tikzpicture}
		\label{c7:fig:traj_straight_kappa}
	}
	~
	\setlength\figureheight{0.22\textwidth}
	\setlength\figurewidth{0.22\textwidth}
	\subfloat[][The trajectories for the joint angles where the stars highlight the initial states.]{
		\begin{tikzpicture}
		\node[anchor=south west] (myplot) at (0,0) {
			\input{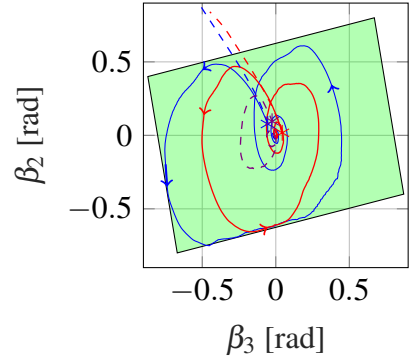}
		};
		\end{tikzpicture}
		\label{c7:fig:traj_straight_beta23}
	}	
	\caption{Experimental results from backward tracking a straight nominal path from three different initial states using the MPC controller (solid lines) and the LQ controller (dashed lines).} 
	\label{c7:fig:traj_straigt}
\end{figure}

\section{Results}
\label{c7:sec:results}
The proposed model predictive path-following controller has been implemented on a modified version of a Scania G580 6x4 tractor that is shown in Figure~\ref{c7:fig:truck_scania}. The tractor is equipped with a servo motor for automated control of the steering column and additional computation power compared to its commercially available version. The tractor is also equipped with a localization system and a rear view LIDAR sensor as illustrated in Figure~\ref{c7:fig:LIDAR_FOV_illustration}. Neither the semitrailer nor the dolly is equipped with any sensors that are used by the system. The vehicle parameters are listed in Table~\ref{c7:tab:vehicle_parameters} and for more details of the test platform, including the used LIDAR-based estimation solution, the reader is referred to~\cite{LjungqvistJFR2019}. 

The MPC controller is implemented in \texttt{C++} and the QP solver qpOASES~\cite{qpOASES} is used to solve~\eqref{c7:MPC_problem} at each sampling instance, where the design parameters for the controller are listed in Table~\ref{c7:tab:design_parameters}. The performance of the MPC controller is evaluated in a set of real-world experiments of backward tracking a straight and a figure-eight nominal path. The performance of the MPC controller is benchmarked with the LQ controller presented in~\cite{LjungqvistJFR2019}, where the LQ feedback gain $K$ is computed by solving the DARE in~\eqref{c7:eq:DARE} using the same weight matrix $Q$ that is used by the MPC controller.

The first set of experiments involve backward tracking ($v=-1$ m/s) of a straight nominal path aligned with the $x$-axis. In the experiments, the initial state $\tilde x(0)$ is perturbed (see Figure~\ref{c7:fig:traj_straight_xy}) to evaluated how the controllers handle disturbance rejection while satisfying the constraint on the joint angles. Experiment 1 involves only a lateral error $\tilde z_3(0) = 5.6$ m, Experiment 2 a heading error \mbox{$\tilde \theta_3(0) =-0.8$ rad} and a small lateral error $\tilde z_3(0) = -1.2$ m, and Experiment 3 includes both a lateral $\tilde z_3(0) = -4.1$ m and a heading error $\tilde \theta_3(0) = -0.42$ rad. The experimental results are presented in Figure~\ref{c7:fig:traj_straigt}. 
In Experiment 1 and 2, the LQ controller is not able to stabilize the vehicle. This is because the LQ controller saturates the tractor's curvature (red and blue dashed lines in Figure~\ref{c7:fig:traj_straight_kappa}) and since the system is open-loop unstable jackknifing occurs almost instantly.
This can be seen in Figure~\ref{c7:fig:traj_straight_beta23} where $\beta_2(\cdot)$ and $\beta_3(\cdot)$ are plotted. As a comparison, the MPC controller is able to make the system converge to the nominal path while satisfying the constraints on the joint angles. In experiment 3, both controllers are able to stabilize the vehicle around the nominal path. The reason for this can be seen in Figure~\ref{c7:fig:traj_straight_kappa} where the curvature of the tractor is plotted. The results shown that the feedback computed by the LQ controller (purple dashed line) is not saturating the tractor's curvature, as opposed to Experiment 1 and 2. 

\begin{figure}[t!]
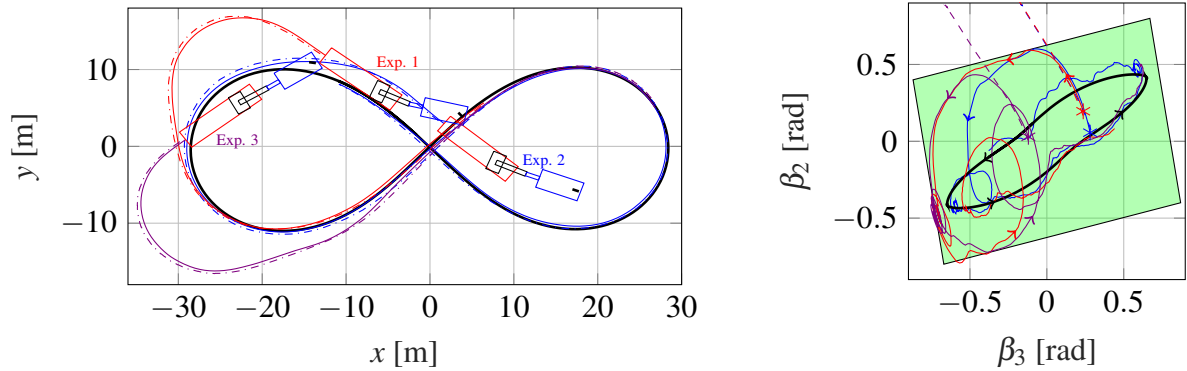

	\centering
	\setlength\figureheight{0.23\textwidth}
	\setlength\figurewidth{0.46\textwidth}
	\captionsetup[subfloat]{captionskip=-2pt}
	\subfloat[][The paths taken by the axle of the semitrailer $(x_3(\cdot),y_3(\cdot))$, the nominal path for $(x_{3r}(\cdot),y_{3r}(\cdot))$ (black solid line), and the measured path of the axle of the semitrailer by the external RTK-GPS (dashed-dotted line).]{
		\begin{tikzpicture}
		\node[anchor=south west] (myplot) at (0,0) {
			\input{path_xy_eight.tex}
		};
		\begin{scope}[x={(myplot.south east)}, y={(myplot.north west)}]
		\node[text=red] at (0.56,0.83) {\tiny Exp. 1};
		\node[text=blue] at (0.76,0.58) {\tiny Exp. 2};
		\node[text=red!50!blue] at (0.34,0.63) {\tiny Exp. 3};
		\end{scope}
		\end{tikzpicture}
		\label{c7:fig:traj_eight_xy}
	}
	~
	\setlength\figureheight{0.23\textwidth}
	\setlength\figurewidth{0.23\textwidth}
	\subfloat[][The trajectories for the joint angles where the stars highlight the initial states, and the black solid line is the nominal path for $(\beta_{3r}(\cdot),\beta_{2r}(\cdot))$.]{
		\begin{tikzpicture}
		\node[anchor=south west] (myplot) at (0,0) {
			\input{trajectory_beta23_eight.tex}
		};
		\end{tikzpicture}
		\label{c7:fig:traj_eight_beta23}
	}
	\caption{Experimental results from backward tracking an figure-eight nominal path from three different initial states using the MPC controller (solid lines) and the LQ controller (dashed lines).} 
	\label{c7:fig:traj_eight}
\end{figure}

The second set of experiments involve backward tracking ($v=-1$ m/s) of a nominal path in the shape of a figure-eight in $(x_{3r}(\cdot),y_{3r}(\cdot))$. Also in this set of experiments, the initial state $\tilde x(0)$ is perturbed (see Figure~\ref{c7:fig:traj_eight_xy}) to compare the performance of the controllers. The experimental results are presented in Figure~\ref{c7:fig:traj_eight}. In all three experiments, the LQ controller fails to stabilize the vehicle and jackknifing occurs almost instantly (see Figure~\ref{c7:fig:traj_eight_beta23}). On the contrary, the MPC controller is able to stabilize the vehicle around the figure-eight nominal path in all three experiments while satisfying the constraints on the joint angles during the majority of the maneuvers. Since the position of the semitrailer only is  estimated, an external RTK-GPS (real-time kinematic GPS) is mounted on the semitrailer's axle to validate the system's performance. As can be seen in Figure~\ref{c7:fig:traj_eight_xy}, even though the maneuvers are quite advanced the observer is able to track the position of the semitrailer's axle. One important reason for this is that the MPC controller is constrained to control the vehicle such that its joint angles remain in the region where high-accuracy state estimates can be computed by the used LIDAR-based estimation solution. As a final note, the average computation time for the MPC controller during the experiments was approximately $5$ ms which is far less than the MPC controller's sampling time of $50$ ms.

\section{Conclusions and future work}
\label{c7:sec:conclusions}
A model predictive path-following controller for a G2T with a car-like tractor is proposed for the case when the nominal path contains full state and control information. Jackknife preventing and sensing limiting constraints on the joint angles are modeled and included in the MPC formulation. The performance of the proposed model predictive path-following controller in terms of suppressing disturbances and satisfying the constraints on the joint angles is evaluated in real-world experiments where it is benchmarked, and shown to outperform, a previously proposed control solutions. As future work, we would like to investigate if there exist other systems with advanced sensors with similar sensing limitations that need to be rigorously respected at the level of control.

\bibliographystyle{abbrv}
\bibliography{root}

\end{document}